\pdfoutput=1

\documentclass[11pt]{article}

\usepackage[final]{acl}
\usepackage{times}
\usepackage{latexsym}
\usepackage{booktabs} 
\usepackage{xcolor} 
\usepackage[table]{xcolor}
\usepackage{colortbl} 

\usepackage{fontawesome}
\usepackage{multirow}
\usepackage{pifont}
\usepackage{amsmath}
\definecolor{darkgreen}{RGB}{0,200,0} 
\definecolor{darkred}{RGB}{200,0,0} 
\usepackage[T1]{fontenc}

\usepackage[utf8]{inputenc}

\usepackage{microtype}

\usepackage{inconsolata}

\usepackage{graphicx}

\title{Conan-Embedding-v2: Training an LLM from Scratch for Text Embeddings}

\author{
 \textbf{Shiyu Li\textsuperscript{1}},
 \textbf{Yang Tang\textsuperscript{1}},
 \textbf{Ruijie Liu\textsuperscript{1}},
 \textbf{Shi-Zhe Chen\textsuperscript{1}},
 \textbf{Xi Chen\textsuperscript{1}\thanks{Corresponding author.}},
\\
\\
  \textsuperscript{1}Basic Algorithm Center, PCG, Tencent \\
  \texttt{\{shyuli, ethanntang, jackrjliu, shizhechen, jasonxchen\}@tencent.com} \\
}

\begin{document}
\maketitle
\begin{abstract}

Large language models (LLMs) have recently demonstrated excellent performance in text embedding tasks. 
Previous work usually use LoRA to fine-tune existing LLMs, which are limited by the data and training gap between LLMs and embedding models.
In this work, we introduce Conan-embedding-v2, a new 1.4B-parameter LLM trained from scratch and fine-tuned as a text embedder.
First, we add news data and multilingual pairs for LLM pretraining to bridge the data gap. 
Based on this, we propose a cross-lingual retrieval dataset that enables the LLM to better integrate embeddings across different languages. 
Second, whereas LLMs use a causal mask with token-level loss, embedding models use a bidirectional mask with sentence-level loss. This training gap makes full fine-tuning less effective than LoRA.
We introduce a soft-masking mechanism to gradually transition between these two types of masks, enabling the model to learn more comprehensive representations.
Based on this, we propose a dynamic hard negative mining method that exposes the model to more difficult negative examples throughout the training process.
Being intuitive and effective, with only approximately 1.4B parameters, Conan-embedding-v2 achieves SOTA performance on both the Massive Text Embedding Benchmark (MTEB) and Chinese MTEB (May 19, 2025).
\end{abstract}

\section{Introduction}
Text embedding maps words, sentences, or documents into a high-dimensional continuous space, allowing similar texts to have closer vector representations~\cite{Mikolov_Sutskever_Chen_Corrado_Dean_2013,Karpukhin_Oguz_Min_Lewis_Wu_Edunov_Chen_Yih_2020}. This representation not only elevates the operability of text data, but also significantly improves performance in various downstream tasks~\cite{Jacob2018bert,radford2018improving,reimers2019sentence}.
With the rapid development of large language models, LLM-based embedding models~\cite{wang2023improving,li2023towards,wang2024multilingual} have played a crucial role in text representation and information retrieval tasks. 

\begin{figure}[t]
\begin{center}
\centerline{\includegraphics[width=\columnwidth]{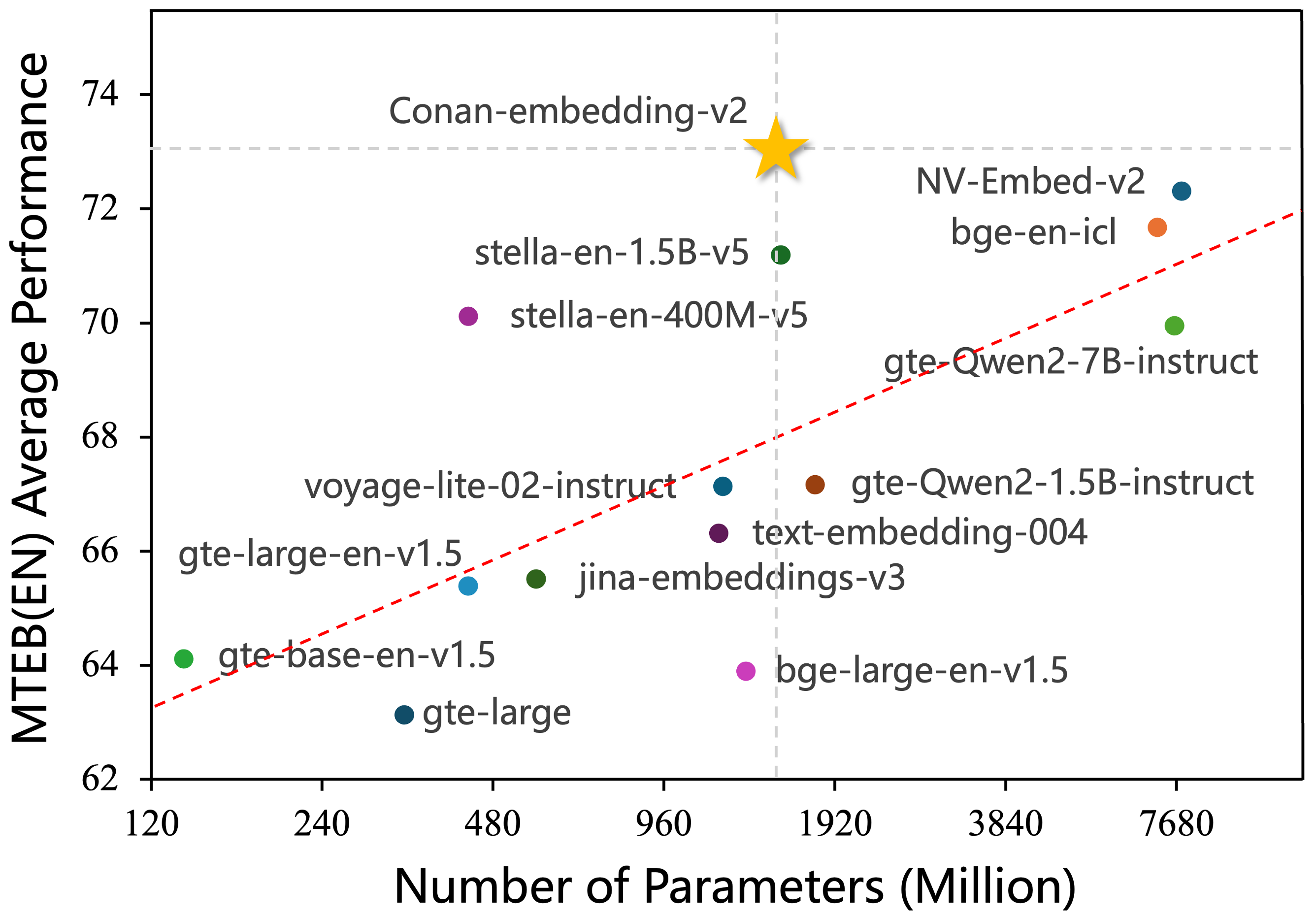}}
\caption{\textbf{Comparison between Conan-embedding-v2 and other embedding models on MTEB English benchmark (May 19, 2025)}~\cite{muennighoff2022mteb}. This benchmark evaluates model performance across seven tasks: classification, clustering, pair classification, reranking, retrieval, semantic textual similarity, and summarization. The red dashed line depicts the logarithmic trendline fitted to the performance data of all the baseline models, excluding Conan-embedding-v2.}
\label{fig:scatter_mteb}
\end{center}
\vskip -0.4in
\end{figure}
However, previous work with LLMs usually starts with the pretrained Mistral-7B~\cite{jiang2023mistral} and LoRA~\cite{hu2021lora} to fine-tune the embedding models. This approach may be constrained by the disparities in the training data and process between LLMs and embedding models. 
First, it relies on the capabilities of the base LLMs, and there is a gap between the corpora used to train the base LLMs and the data required for embedding training.
Moreover, the training paradigms for LLMs and embedding models are fundamentally different. 
LLMs are trained to predict the next token, whereas embedding models need to generate an embedding vector based on the entire query or candidate sentence. 
This training gap makes full fine-tuning less effective than LoRA, and the improvements through LoRA have inherent limitations~\cite{biderman2024lora}.

To address the above challenges, we propose Conan-embedding-v2, a new LLM trained from scratch and finetuned as a text embedder, which extends the BERT-based conan-v1 \cite{li2024conan} in both training data and methodology.
First, to bridge the data gap, Conan-embedding-v2 combines pretraining on extensive news data with fine-tuning on specialized embedding corpora during LLM training.
Second, to address the training gap, we have developed a \textbf{soft mask} mechanism that facilitates a gradual transition from causal masking to bidirectional masking, allowing the rank of the mask to gradually decrease.
This enables the model to learn more comprehensive feature representations during the early stages of training. 
Specifically, as LLMs are no longer constrained by the corpus of the backbone, we introduce a novel \textbf{cross-lingual retrieval dataset} that enables bidirectional search between 26 languages. This allows the model to integrate embeddings across diverse linguistic systems. 
Moreover, since LLMs are no longer constrained by LoRA, we present a \textbf{dynamic hard negative mining} that keep the high value of negative samples throughout the training process. 

As shown in Figure~\ref{fig:scatter_mteb}, Conan-embedding-v2 demonstrates SOTA performance, outperforming both BERT-based and LLM-based methods, while maintaining an efficient model architecture with optimized size. Our key contributions can be summarized as follows:

\begin{itemize}
  \item We propose Conan-embedding-v2, a new LLM trained from scratch and finetuned as a text embedder to address the data and training gaps between LLMs and embedding models.

  \item We introduce a novel cross-lingual retrieval dataset that enables bidirectional search between 26 languages, improving the integration of multilingual embeddings.

  \item We conduct empirical evaluations, demonstrating that our method achieves SOTA performance on both English and Chinese MTEB benchmarks, while maintaining a reasonable model size and inference speed.
  
\end{itemize}

\section{Related Work}
\subsection{LLM-based Embedding Models}

Recent progress in LLMs has significantly advanced the development of text embedding models, enabling more efficient and versatile representations. By fine-tuning pretrained LLMs on the synthetic data, ~\cite{wang2023improving} achieved outstanding performance with few training steps. The findings of this research confirmed that leveraging LLMs for embeddings proved efficient and effective. Researchers have proposed diverse approaches to enhance LLM-based text embedders from multiple perspectives. NV-Embed~\cite{lee2024nv} improved representation capability through introducing latent attention layers and removing causal attention encoding. bge-en-icl~\cite{li2024making} utilized a few-shot learning approach to generate high-quality text embeddings by taking advantage of the in-context learning ability in LLMs. NV-Retriever~\cite{moreira2024nv} introduced a mining approach using positive relevance scores to eliminate false negatives, improving training efficiency and retrieval accuracy. mE5~\cite{wang2024multilingual} and M3-Embedding~\cite{chen-etal-2024-m3} focused on multilingual text embedding. The above research significantly improved the performance of LLM-based text embedding.

\subsection{Cross-lingual Information Retrieval}
While LLM-based embedding models have shown remarkable progress, their application in cross-lingual information retrieval (CLIR) presents unique challenges and opportunities~\cite{hammerl2024understanding}. Traditional CLIR methods struggle to support multiple languages, maintain computational efficiency, and achieve high retrieval performance simultaneously. Recent advances have demonstrated promising developments. Multilingual text embedding approaches, such as M3-Embedding~\cite{chen-etal-2024-m3} and mE5~\cite{wang2024multilingual}, have shown remarkable capabilities in handling multiple languages while maintaining computational efficiency through contrastive learning and knowledge distillation techniques. Additionally, LECCR~\cite{wang2024multimodal} has begun incorporating multimodal LLMs to bridge the semantic gap between different modalities and languages, resulting in significant improvements in cross-lingual cross-modal retrieval tasks. To address the challenges of low-resource languages, recent studies~\cite{miao2024enhancing,litschko2024cross} have proposed innovative solutions using word alignment and dialect-specific approaches to enhance embedding quality.

\section{Method}\label{sec:method}

\begin{figure*}[t]
\begin{center}
\centerline{\includegraphics[width=2.0\columnwidth]{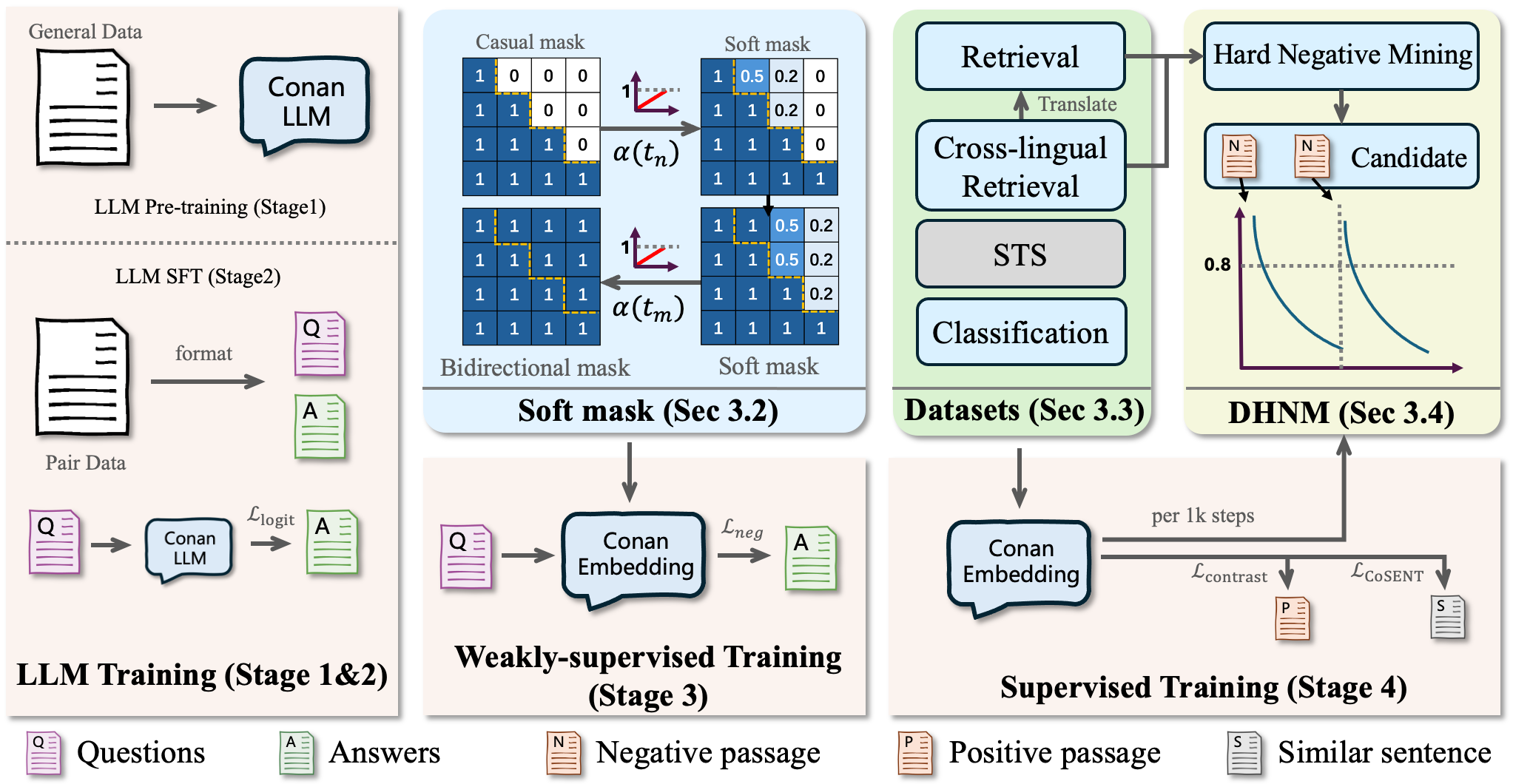}}
\caption{Overview of Conan-embedding-v2: Our approach consists of four stages. During LLM training (stages 1 and 2), we add embedding data to better align the LLM with embedding tasks. In the weakly-supervised training stage, we use the same pairs from LLM SFT and apply a soft mask to bridge the gap between LLMs and embedding models. In the supervised training stage, benefiting from LLM training, we incorporate a cross-lingual retrieval dataset and a dynamic hard negative mining approach to improve data diversity and value.}
\label{fig:framework}
\end{center}
\vskip -0.3in
\end{figure*}
\subsection{Overall Pipeline}
Since Conan-embedding-v2 is trained from scratch, the training process is divided into four stages: LLM pre-training, LLM supervised fine-tuning (SFT), embedding weakly-supervised training, and embedding supervised training. Each stage differs in data formats and loss functions.

\subsubsection{LLM Training}\label{sec:llm_train}
To better adapt large language models (LLMs) to embedding tasks, we designed Conan-embedding-v2 with 8 layers and a hidden dimension of 3584, supporting up to 32,768 input tokens. This model, totaling 1.4 billion parameters, offers a higher number of embedding dimensions with fewer parameters. We trained the Conan tokenizer on approximately 400,000 multilingual corpora, resulting in a vocabulary size of 150,000.
As shown in Figure~\ref{fig:framework}, we initially pre-trained the model on approximately 3T tokens of general data, with a emphasis on adding news, question-answer, and web page data.
We employed the standard data filtering methods as described in \cite{cai2024internlm2}.
Following this, we collected approximately 600 million instances of SFT data using the paired data (query-positive), formatted as instruction, input, and output. 
\subsubsection{Embedding Training}
\textbf{Weakly-supervised Training.}
For embedding training, we first implemented weakly-supervised training to allow the model to initially learn the representations for embedding. 
During this stage, we use the same data as in LLM supervised fine-tuning, but with different data formats and loss functions.
Specifically, we provide the instruction and input as the query, and the output as the positive passage. 
To ensure higher data quality, we utilize the gte-Qwen2-7B-instruct model~\cite{li2023towards} for scoring and discard any data with scores below 0.4. 
To efficiently and effectively leverage the pair data, we employ the InfoNCE loss with In-Batch Negative sampling~\cite{gutmann2010noise} during training, the formula is:
\begin{equation}
\mathcal{L}_{neg} = - \sum_{i=1}^N \log \frac{\exp(\text{cos}(x_i, y_i^+))}{\sum_{j=1}^M \exp(\text{cos}(x_i, y_i))}
\end{equation}
$x_i$ represents the query of the positive sample, $y_i^+$represents the passage of the positive sample, $y_i$ represents the passages of other samples in the batch, which are considered as negative samples.

\noindent\textbf{Supervised Training. }
After weakly-supervised training, we perform task-specific fine-tuning for different downstream tasks. 
As shown in Figure~\ref{fig:framework}, we divide the tasks into four categories: retrieval, cross-lingual retrieval, classification and STS (semantic textual similarity). 
The first three tasks include a query, a positive text, and some negative texts, utilizing the classic InfoNCE loss function.
STS task involves distinguishing the similarity between two texts, with the classic loss function being cross-entropy loss. 
According to~\cite{kexuefm-9341} and other works~\cite{Moka2023}, CoSENT loss is slightly better than cross-entropy loss. 
Therefore, we also adopt CoSENT loss to optimize STS task, which is formulated as follows:
\begin{equation}
\mathcal{L}_{cos} = \log \left(1 + \sum\limits_{Order} \exp \frac{\langle x_k, x_l \rangle - \langle x_i, x_j \rangle}{\tau}\right)
\end{equation}
where $Order = \text{sim}(i,j) > \text{sim}(k,l)$, $\text{sim}(k, l)$ is the ground-truth similarity between $x_i$ and $x_j$. $\langle x_k, x_l \rangle$ represents the cosine similarity between $x_k$ and $x_l$. $\tau$ is the scale temperature.



\subsection{Soft Mask}\label{sec:soft_mask}
During the training phase of LLMs, a causal mask is employed to ensure that the current token does not have access to subsequent tokens, which is suitable for token-level language modeling. 
However, embedding training requires a holistic understanding of the sentence, using a bidirectional mask for vector-level modeling. These two types of masks have several key gaps.

First, since the upper triangle of the causal mask is entirely zeros, the attention weights in this region are not used during the forward propagation. 
When switching directly to a bidirectional mask, these weights require a learning process to become effective.
Second, the causal mask is full-rank, providing stronger expressive power, whereas the rank of the bidirectional mask is always one.
If we directly switch from a causal mask to a bidirectional mask during the weakly supervised fine-tuning stage, the training may initially converge quickly due to the low rank but is prone to getting stuck in local minima, making further optimization challenging. 

As illustrated in Figure~\ref{fig:framework}, to address these gaps, we introduce a novel soft mask mechanism. 
Firstly, to address the issue of attention weights, we introduce an $\alpha(t)$ term in the soft mask, where $\alpha(t)$ is our scheduling function, allowing the mask to gradually transition from 0 to 1, enabling the model to progressively update these parameters. $\tau$ is set to the total number of steps for normalization. 
$\alpha(t)$ is defined as follows:
\begin{equation}
\alpha(t) = \frac{t}{\tau}
\end{equation}
Secondly, as weakly supervised training requires learning richer feature representations, we propose a dynamic rank reduction approach. We use $M_{ij}$ to represent the mask matrix. 
We employed a simple method where the values of the first $i$ column of $M_{ij}$ are set to 1, resulting in a rank of $N-i$.
By combining this with our weight adjustment method, the values closer to the beginning transition to 1 more quickly. The formula for the soft mask is as follows:
\begin{equation}
M_{ij}(t) = \begin{cases} 
1 & \text{if } i \geq j \\ 
\min\left(\alpha(t) \times \frac{l}{i}, 1\right) & \text{if } i < j
\end{cases}
\end{equation}
$i<j$ indicates that we are modifying the upper triangular values. $l$ is the training sequence length. We ensure that the maximum value is 1, and the earlier columns reach 1 sooner. 
This not only allows the rank to gradually decrease but also aligns with the trend of reading from front to back, where the weights gradually decrease. 
We will discuss the impact of different $\alpha(t)$ on the results in Appendix~\ref{appendix:softmaskfunc}.

\subsection{Cross-lingual Retrieval Dataset}\label{sec:cross_lingual}
To develop a multilingual LLM, we aim for Conan-embedding-v2 to learn representations across different languages. 
Previous work has primarily focused on fine-tuning using multilingual corpora directly or using parallel corpora where the texts are translations, often overlooking the intrinsic relationships between languages.
To address this issue, we propose a cross-lingual retrieval dataset (CLR), which integrates representations across different languages through cross-lingual search, thereby narrowing the representation gap between them.

We start with existing retrieval datasets and extend them to support cross-lingual retrieval. To reduce the workload, we only translate the query portion of the datasets using Qwen2.5-7B~\cite{qwen2.5}. For instance, we translate the queries in MSMARCO~\cite{nguyen2016ms} (an English retrieval task) subset to Chinese to enable Chinese-to-English retrieval. Similarly, we apply this approach to other tasks, translating queries to support cross-retrieval among 26 languages, resulting in a total of approximately 10 million pairs.
\begin{figure}[h]
    \centering
    \includegraphics[width=1.0\linewidth]{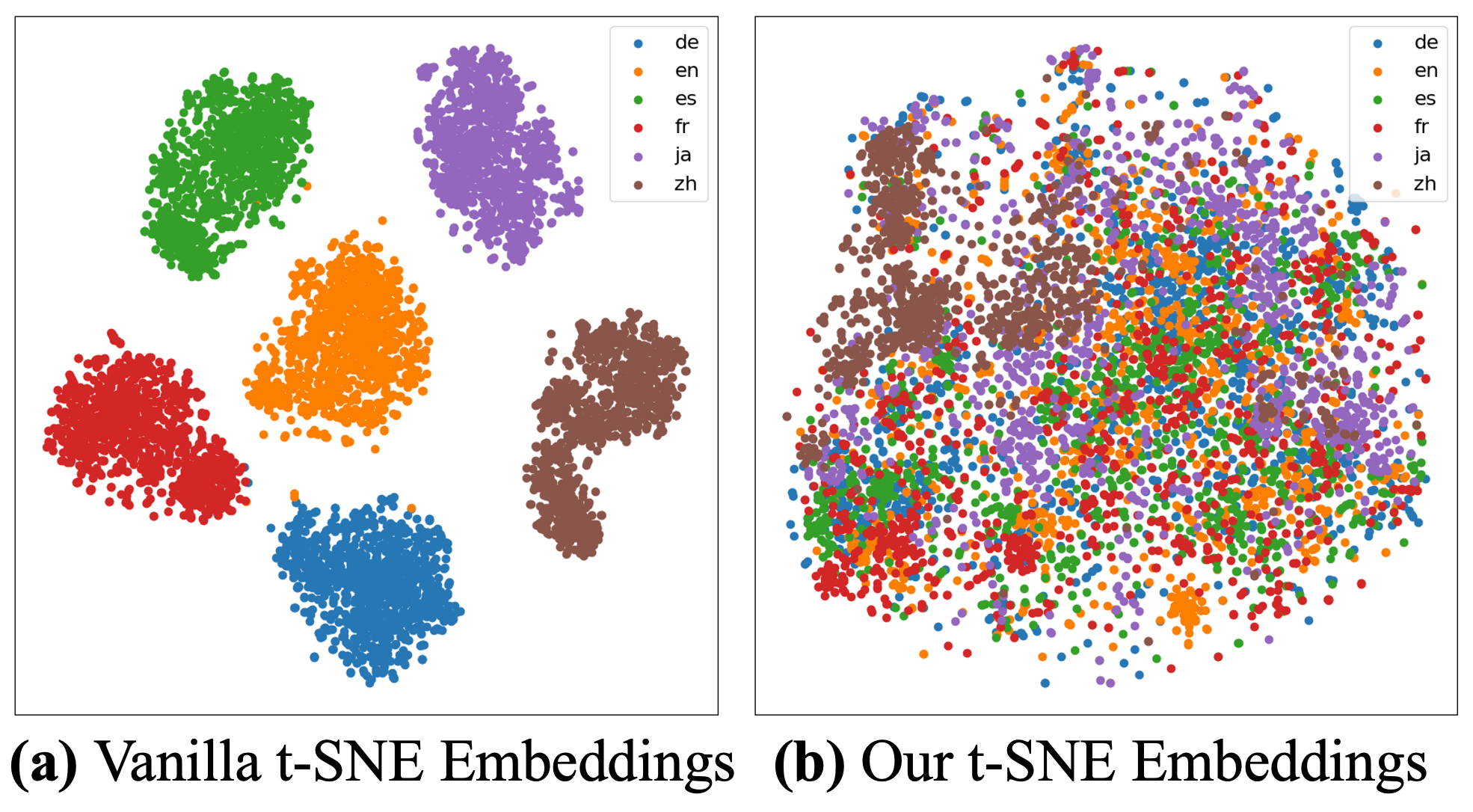}
    \caption{Embedding distribution before and after training on cross-lingual retrieval dataset.}
    \label{fig:CLIR_embed}
    \vskip -0.0 in
\end{figure}

To provide a more intuitive representation of the embeddings, we conducted a comparative analysis of the embedding distribution. 
We utilized the Multilingual Amazon Reviews Corpus~\cite{marc_reviews}, which is not included in our cross-lingual retrieval dataset. 
This corpus includes reviews in English, Japanese, German, French, Chinese, and Spanish. 
For each language, we sampled 1000 sentences from the test set. 
As shown in Figure~\ref{fig:CLIR_embed}, the vanilla method represents our model without the CLR dataset included. 
The embeddings for six different languages were distinctly clustered, each occupying a separate region in the distribution space. 
In contrast, our model Conan-embedding-v2 successfully integrated the embeddings of all languages into a unified distribution, demonstrating its effectiveness in creating a more cohesive multilingual representation.

\subsection{Dynamic Hard Negative Mining}
Previous work has primarily focused on hard negative mining during the data preprocessing stage using existing embedding models, resulting in fixed hard negatives. 
However, the hard negatives identified by other models may differ from those identified by the model being trained. 
Furthermore, as training progresses and model weights are updated, the score of hard negatives corresponding to the current weights change. 
Hard negatives mined during the preprocessing stage may become less challenging after several training iterations.

Based on this insight, we propose a dynamic hard negative mining (DHNM) method in conan-v1 \cite{li2024conan}. DHNM dynamically detects the difficulty of the current sample during the training process and replaces the sample based on its difficulty. We use scores to represent the difficulty level, the formula is as follows:
\begin{equation}
S= \text{cos}\langle f(q), f(p)\rangle
\end{equation}
$S$ represents the cosine score, $f_{k}(q)$ is query embedding, and $f(p)$ is hard negative embedding.

Unlike the replacement criteria in v1, in this paper, if the absolute value of the score is less than 0.4 at the initial step, it will also be discarded. The current detection formula is:
\begin{equation}
\mathbf{N}_{i} = \begin{cases}
\mathbf{N}_{i+1} & (S_{0} < 0.4) \\
\mathbf{N}_{i+1} & 1.2\cdot S_{i} {\scriptstyle <} S_{0} \;  \& \; S_{i} {\scriptstyle <} 0.7 \\
\mathbf{N}_{i} & \text{otherwise}
\end{cases}
\end{equation}

${N}_{i}$ denotes the $i$-th hard negative sample, with $S_{i,0}$ as its initial score and $S_{i}$ as its current score. If the score multiplied by 1.2 is less than the initial score and the absolute value of the score is less than 0.7, we consider the negative example no longer difficult. We replace it with a new hard negative $N_{i+1}$ from the hard negative pool.

Additionally, in v1, the check is performed every 1k steps. In this paper, we leverage the fact that the similarity score between the query and each hard negative is already computed as part of the loss calculation. During each loss computation, we add a lightweight check to cache the current score of each hard negative and determine whether it is still sufficiently challenging for the model. If a hard negative’s score indicates that it is no longer difficult, we mark it for replacement. In the next training step, we replace this negative with a new hard negative sampled from the candidate pool. This process ensures that the set of hard negatives remains up-to-date and challenging throughout training, without introducing additional computational overhead.

\section{Experiments}
 

\begin{table*}[h!]
    \centering
    \resizebox{1.00\linewidth}{!}{
    \begin{tabular}{c|l|cccccccc}
\toprule
\multirow{2}{*}{\textbf{Languages}}
& \textbf{Embedding Task} & \textbf{Class.} & \textbf{Clust} & \textbf{PairClass} & \textbf{Rerank} & \textbf{Retri} & \textbf{STS} & \textbf{Summ.} & \textbf{Avg.} \\
& Metric & Acc. & V-Meas. & AP & MAP & nDCG@10  & Spear. & Spear. & \\
\midrule
\multirow{7}{*}{\textbf{English}}
& e5-mistral-7b-instruct & 79.85 & 51.44 & 88.42 & 49.78 & 57.62 & 84.32 & 36.57 & 67.97 \\
& stella-en-1.5B-v5 & 89.38 & 57.06 & 88.02 & 50.19 & 52.42 & 83.27 & 36.91 & 69.43 \\
& NV-Embed-v2 & 87.19 & 47.66 & 88.69 & 49.61 & 62.84 & 83.82 & 35.21& 69.81  \\
& gte-Qwen2-7B-instruct & 88.52 & 58.97 & 85.9 & \textbf{50.47} & 58.09 & 82.69 & 35.74 & 70.72 \\
& jasper-en-v1 & 90.27 & \textbf{60.52} & 88.14 & 50 & 56.05 & 84.37 & 37.19 & 71.41 \\
& gemini-embedding-exp-03-07 & 90.05 & 59.39 & 87.70 & 48.59 & 64.35 & \textbf{85.29} & \textbf{38.28} & 73.30 \\
& \cellcolor{cyan!10}\textbf{Conan-embedding-v2} & \cellcolor{cyan!10}\textbf{90.98} & \cellcolor{cyan!10}59.96 & \cellcolor{cyan!10}\textbf{92.35} & \cellcolor{cyan!10}49.07 & \cellcolor{cyan!10}\textbf{66.24} & \cellcolor{cyan!10}85.12 & \cellcolor{cyan!10}35.48 & \cellcolor{cyan!10}\textbf{73.52} \\
\midrule
\multirow{8}{*}{\textbf{Chinese}}
& e5-mistral-7b-instruct      & 72.96 & 52.30 & 66.31 & 61.38 & 61.75 & 48.34 & - & 59.92 \\
& gte-Qwen2-1.5B-instruct     & 72.53 & 54.61 & 79.50 & 68.21 & 71.86 & 60.05 & - & 67.12 \\
& bge-multilingual-gemma2     & 75.31 & 59.30 & 79.30 & 68.28 & 73.73 & 55.19 & - & 67.64 \\
& gte-Qwen2-7B-instruct       & 75.77 & 66.06 & 81.16 & 69.24 & 75.70 & 65.20 & - & 71.62 \\
& xiaobu-embedding-v2         & 76.53 & 65.17 & 85.94 & 72.58 & 76.49 & 64.18 & - & 72.36 \\
& Conan-embedding-v1          & 76.77 & 66.33 & 85.68 & 72.76 & 76.67 & 63.67 & - & 72.50 \\
& retrieve-zh-v1            & \textbf{76.88} & 66.50 & 85.98 & 72.86 & 76.97 & 63.92 & - & 72.71 \\
& \cellcolor{cyan!10}\textbf{Conan-embedding-v2} & \cellcolor{cyan!10}76.47 & \cellcolor{cyan!10}\textbf{68.84} & \cellcolor{cyan!10}\textbf{92.44} & \cellcolor{cyan!10}\textbf{74.41} & \cellcolor{cyan!10}\textbf{78.31} & \cellcolor{cyan!10}\textbf{65.48} & \cellcolor{cyan!10}- & \cellcolor{cyan!10}\textbf{74.24} \\
\bottomrule
\end{tabular}}
\caption{Results for MTEB in English and Chinese.} 
\label{table:mteb_en_zh}
\end{table*}
\subsection{Training Data}\label{sec:train_data}
To achieve the multilingual capability of Conan-embedding-v2, we collected large and diverse data for weakly supervised pre-training and embedding fine-tuning. For \textbf{weakly supervised pre-training}, we primarily gathered title-content pairs from news articles and websites, specifically from CC-News~\cite{Hamborg_Meuschke_Breitinger_Gipp_2017}, mC4~\cite{Karpukhin_Oguz_Min_Lewis_Wu_Edunov_Chen_Yih_2020}, Wikipedia and Chinese Corpora Internet~\cite{baai-cci}.
To ensure data integrity, we applied the Data-Juicer~\cite{djv1} tool for systematic removal of low-quality samples, redundant duplicates, and potentially harmful content.

\noindent\textbf{Embedding supervised training.} For both Chinese and English, we compiled datasets for five different tasks: retrieval, reranking, classification, clustering semantic textual similarity (STS). We ensured that any training data matching the MTEB evaluation set was filtered out. Detailed data usage is provided in Appendix~\ref{sec:data_details}.

\subsection{Model Architecture}\label{sec:model_details}
As demonstrated in \cite{kaplan2020scaling}, under a fixed parameter budget, increasing the number of transformer layers beyond seven results in test loss that remains almost constant. 
Motivated by this observation, we strategically selected eight transformer layers, thereby allocating more parameters to the hidden size and the number of attention heads. This design choice aims to maximize the model’s theoretical representational capacity within the given parameter constraints. 
Consequently, although our model contains only 1.4 billion parameters, it retains the same hidden size as gte-Qwen2-7B-instruct \cite{li2023towards} (3584 dimensions with 28 hidden layers and 28 attention heads). 
In addition, our model is configured with 32 attention heads and 8 key-value heads using GQA optimization, 8192 intermediate dimensions for the feed-forward network layers, a maximum context window of 32,768 tokens, and a vocabulary size of 150,000 tokens.

\subsection{MTEB Results}
In this section, we present the experimental results of our method on the MTEB English and MTEB Chinese benchmarks, and compare it with other SOTA methods. 

 \begin{table*}[h!]
    \centering
    \resizebox{1.00\linewidth}{!}{
    \begin{tabular}{l|cccccccccc}
\toprule
\textbf{Embedding Task}& \textbf{Zero-shot} & \textbf{Class.} & \textbf{Clust.} & \textbf{PairClass.} & \textbf{Rerank.} & \textbf{Retri.} & \textbf{STS} & \textbf{Summ.} & \textbf{Avg.} \\
Metric & & Acc. & V-Meas. & AP & MAP & nDCG@10  & Spear. & Spear. & \\
\midrule 	
bge-large-en-v1.5 & 100\% & 78.34 & 48.01 & 87.13 & 48.26 & 55.44 & 82.79 & 33.13 & 65.89   \\
multilingual-e5-large-instruct & 95\% & 75.54 & 49.89 & 86.24 & 48.74 & 53.47 & 84.72 & 29.89 & 65.53  \\
GIST-Embedding-v0 & 80\%  & 78.16 & 48.50 & 86.33 & 47.52 & 53.59 & 83.35 & 32.32 & 65.50 \\
UAE-Large-v1 & 100\%  & 79.08 & 47.86 & 87.25 & 48.35 & 55.91 & 84.37 & 30.13 & 66.40 \\
mxbai-embed-large-v1 & 100\% & 79.10 & 47.48 & 87.20 & 48.05 & 55.40 & 84.42 & 32.63 & 66.26 \\
GritLM-7B & 95\% & 81.25 & 50.82 & 87.29 & 49.59 & 54.95 & 83.03 & 35.65 & 67.07 \\
e5-mistral-7b-instruct & 95\% & 79.85 & 51.44 & 88.42 & 49.78 & 57.62 & 84.32 & 36.57 & 67.97  \\
text-embedding-005 & 95\%  & 86.03 & 51.91 & 87.62 & 48.84 & 58.77 & \textbf{85.18} & 35.05 & 69.60 \\
SFR-Embedding-Mistral & 85\% & 80.47 & 54.93 & 88.59 & \textbf{50.15} & 59.33 & 84.77 & 36.32 & 69.31  \\
Linq-Embed-Mistral & 95\%  & 83.00 & 54.07 & 88.44 & 49.44 & 60.14 & 84.69 & \textbf{37.26} & 69.80 \\
\rowcolor{cyan!10}\textbf{Conan-embedding-v2} & 95\% & \textbf{88.35} & \textbf{57.34} & \textbf{90.97} & 
47.21 & \textbf{63.84} & 83.77 & 35.20 & \textbf{71.43} \\
\bottomrule
\end{tabular}}
\caption{Zero-shot MTEB results in English. }
\label{table:mteb_zero_shot}
\end{table*}

\noindent\textbf{Results for MTEB in English and Chinese.}
Table~\ref{table:mteb_en_zh} provides a detailed comparison of our method's performance on the MTEB English (classic) benchmark and the MTEB Chinese benchmark. 
The English benchmark additionally includes a summary (summ.) task, which is similar to the STS task. Both tasks measure sentence similarity using Spearman's correlation coefficient. 
Conan-embedding-v2 achieves SOTA results in both English and Chinese benchmark, excelling in classification (91.11 in English, and 76.8 in Chinese) and reranking (51.49 in English, and 73.69 in Chinese) through multiple training strategies. This is consistent with the results observed in the Multilingual benchmark. 
Notably, Conan-embedding-v2 performs slightly worse than other models on the STS tasks, which may be due to the lower proportion of STS data compared to other training task.

\noindent\textbf{Results for MTEB in English in zero-shot.} To validate the effectiveness and generalization ability of our proposed method, we followed the data selection strategy of e5-mistral-7b-instruct \cite{wang2023improving} and used only a small portion of the MTEB training dataset for zero-shot training. The selected datasets include MSMARCO~\cite{nguyen2016ms}, NQ~\cite{Kwiatkowski_Palomaki_Redfield_Collins_Parikh_Alberti_Epstein_Polosukhin_Devlin_Lee_e2019}, XQuADRetrieval~\cite{rajpurkar-etal-2016-squad}, FEVER~\cite{Thorne_Vlachos_Christodoulopoulos_Mittal_2018}, HotpotQA~\cite{Yang_Qi_Zhang_Bengio_Cohen_Salakhutdinov_Manning_2018}, MIRACLRetrieval~\cite{zhang2023miracl}, and MrTidyRetrieval~\cite{Zhang_Ma_Shi_Lin_2021}. 
Table~\ref{table:mteb_zero_shot} summarizes our model’s performance on the MTEB English benchmark in the zero-shot setting. Compared to Linq-Embed-Mistral (7B), our Conan-embedding-v2 (1.4B) achieves a significant improvement. These findings demonstrate the strong zero-shot performance and efficiency of our approach, even with significantly smaller models, validating our innovations in training from scratch and employing novel soft mask techniques to address representational gaps.

\subsection{MKQA Benchmark}
To evaluate cross-lingual retrieval performance, we conducted comprehensive experiments using the Multilingual Knowledge Questions \& Answers (MKQA) benchmark proposed by \cite{longpre2021mkqa}. 
This benchmark provides professionally translated queries and contains 10,000 question-answer pairs from NQ~\cite{Kwiatkowski_Palomaki_Redfield_Collins_Parikh_Alberti_Epstein_Polosukhin_Devlin_Lee_e2019}, aligned across 26 typologically diverse languages (260k question-answer pairs in total).

Following previous works~\cite{izacard2021unsupervised, chen-etal-2024-m3}, we conducted retrieval in NQ~\cite{Kwiatkowski_Palomaki_Redfield_Collins_Parikh_Alberti_Epstein_Polosukhin_Devlin_Lee_e2019} for a given question in a specific language and evaluate whether the English passage is present in the retrieved documents. 
For multilingual models, we computed nDCG@10 and Recall@k (k=20,100) across all 25 target languages to assess both ranking precision and answer coverage. We present the performance details for all languages in the Appendix~\ref{appendix:mkqa}. 

As shown in Table~\ref{table:CLIR}, our proposed Conan-embedding-v2 achieves SOTA performance, outperforming all of the baseline models across all metrics. Notably, it achieves significant improvements of +3.6\% R@20 and +5.7\% nDCG@10 over the strongest baseline (M3-Embedding), demonstrating superior cross-lingual alignment capability.

\subsection{Ablation Study}
We systematically evaluated the contributions of individual components in our framework through ablation experiments (Table~\ref{table:ablation}). 
The isolated Cross-lingual Retrieval task objective (Row 2) improves multilingual performance to 62.69\% (+1.96\% Multi over SM-only) while maintaining stable single-language scores, demonstrating its targeted capability for cross-lingual representation refinement.
Using only Dynamic Hard Negative Mining (Row 3) yields the best language-specific results among single components (71.50\% Eng/72.09\% Zh), confirming its effectiveness in distinguishing fine-grained semantic boundaries through adaptive negative sampling.
The combination of SM+CLR (Row 4) produces a significant multilingual performance leap to 64.45\% (+3.56\% over SM-only), while SM+DHNM (Row 5) achieves peak language-specific scores before full integration. 
However, both partial combinations reveal an accuracy tradeoff between multilingual and single-language task.
Our complete framework with all components (bottom row) resolves this tradeoff by synergistically combining SM’s initialization stability, CLR’s cross-lingual alignment, and DHNM’s discriminative training, achieving SOTA performance across all tasks. 
These results validate the synergistic effects of Conan-embedding-v2 components in enhancing the model's overall capabilities.

\subsection{Analysis}

\begin{table*}[t]
\centering
\minipage{0.45\linewidth}
    \centering
    \begin{small}
    \begin{tabular}{lcccc}
    \toprule
     \textbf{Model} & \textbf{R@20} & \textbf{R@100} & \textbf{nDCG@10}\\
    \midrule
    BM25 & 28.1 & 39.9 & 25.4 \\
    mContriever & 56.3 & 67.9 & 44.9 \\
    text-embedding-v3 & 62.1 & 69.5 & 48.1 \\
    e5-mistral & 62.4 & 70.1 & 47.5 \\
    M3-Embedding & 68.8 & 75.5 & 53.2 \\
    \midrule
    \rowcolor{cyan!10}
    \textbf{Conan-embedding-v2} & \textbf{72.5} & \textbf{80.2} & \textbf{59.1} \\
    \bottomrule
    \end{tabular}
    \caption{Results of cross-lingual retrieval performance on MKQA benchmark.}
    \label{table:CLIR}
    \end{small}
    \vspace{-10pt}
\endminipage
\hspace{1cm}
\minipage{0.45\linewidth}
    \centering
    \begin{small}
    \begin{tabular}{ccccccc}
    \toprule
    \textbf{SM} & \textbf{CLR} & \textbf{DHNM} & \textbf{Multi} & \textbf{Eng} & \textbf{Zh} \\
    \midrule
    \textcolor{darkgreen}{\ding{52}} & \textcolor{darkred}{\ding{55}} & \textcolor{darkred}{\ding{55}} & 61.73 & 70.41 &  70.99 \\
    \textcolor{darkred}{\ding{55}} & \textcolor{darkgreen}{\ding{52}} & \textcolor{darkred}{\ding{55}} & 62.69 & 70.94 & 71.41 \\
    \textcolor{darkred}{\ding{55}} & \textcolor{darkred}{\ding{55}} & \textcolor{darkgreen}{\ding{52}} & 61.81 & 71.50 & 72.09 \\
    \textcolor{darkgreen}{\ding{52}} & \textcolor{darkgreen}{\ding{52}} & \textcolor{darkred}{\ding{55}} & \underline{64.45} & 72.14 & 71.79 \\
    \textcolor{darkgreen}{\ding{52}} & \textcolor{darkred}{\ding{55}} & \textcolor{darkgreen}{\ding{52}} & 63.03 & \underline{72.78} & \underline{72.44} \\
    \midrule
    \rowcolor{cyan!10}
    \textcolor{darkgreen}{\ding{52}} & \textcolor{darkgreen}{\ding{52}} & \textcolor{darkgreen}{\ding{52}} & \textbf{65.17} & \textbf{73.52} & \textbf{74.24} \\
    \bottomrule
    \end{tabular}
    \caption{Results of ablation study on MTEB. SM, CLR and DHNM are defined in Sec~\ref{sec:method}.}
    \label{table:ablation}
    \end{small}
    \vspace{-10pt}
\endminipage
\end{table*}

\begin{table*}[h!]
\vskip 0.1in
    \centering\resizebox{0.99\linewidth}{!}{
    \begin{tabular}{l|cccccccccc}
\toprule
\textbf{Model} & \textbf{Model Size (million)} & \textbf{Embedding Dim.} & \textbf{Infer Time (min.)} & \textbf{MRL} & \textbf{Avg.} \\
\midrule
gte-large-en-v1.5 & 335	& 1024 & 1.12 & \textcolor{darkred}{\ding{55}} & 65.89 \\
stella-en-1.5B-v5 & 1543 & 1536 & 5.54 & \textcolor{darkgreen}{\ding{52}} & 69.43 \\
Linq-Embed-Mistral & 6782  & 4096 & 30.61 & \textcolor{darkred}{\ding{55}} &  69.80 \\
NV-Embed-v2 & 7851 & 4096 & 33.58 & \textcolor{darkred}{\ding{55}} & 69.81 \\
gte-Qwen2-7B-instruct & 7613 & 3584 & 31.78 & \textcolor{darkred}{\ding{55}} & 70.72 \\

\textbf{Conan-embedding-v2} & 1503 & 3584 & 5.14 & \textcolor{darkgreen}{\ding{52}} & 73.52 \\
\bottomrule
\end{tabular}}
\caption{Comparison of practical factors of different embedding models.}
\label{table:practical}
\vspace{-5pt}
\end{table*}

\subsubsection{Practical Considerations}
In addition to performance, the practical selection of an embedding model is influenced by many other factors. 
To better demonstrate how our model ensures both efficiency and applicability, we have also highlighted several other important factors.
We select model size, embedding dimension, inference time, and support for Matryoshka Representation Learning (MRL)~\cite{kusupati2022matryoshka}. 
MRL indicates whether the model supports embeddings of different dimensions. 
Inference time is measured in minutes and is based on the results obtained using the English queries from the train set of the Multilingual Amazon Reviews Corpus on a single 910B GPU.
Additionally, we provide the MTEB English benchmark results as a performance reference.

Table~\ref{table:practical} shows a comparison between several representative models and our model.
Conan-embedding-v2 stands out by maintaining a balanced model size of 1503 million parameters and an embedding dimension of 3584. 
Despite its compact size, Conan-embedding-v2 achieves an impressive inference time of just 5.14 minutes, making it one of the fastest models evaluated.
Additionally, Conan-embedding-v2 supports MRL, a capability shared only with stella-en-1.5B-v5. 
However, stella-en-1.5B-v5 has a smaller embedding dimension of 1536 and slightly lower performance, with an average score of 71.19. 
This highlights Conan-embedding-v2's superior efficiency and effectiveness in practical applications.
\subsubsection{Training Gap}

Token-level LLM training loss and sentence-level contrastive loss have fundamentally different optimization landscapes. Full fine-tuning forces an abrupt transition between these paradigms, causing representation collapse \cite{luo2023empirical}. In contrast, LoRA updates only a small subset of parameters, providing a smoother optimization path \cite{zhang2024mgte}. 
Table \ref{table:train_gap} compares MTEB-EN results using different methods on Conan-embedding-v2. The results confirm the findings in \cite{zhang2024mgte}. However, with soft mask applied, higher LoRA ranks consistently yield better results. This demonstrates that soft mask effectively bridges the gap between LLM's generative training and contrastive learning objectives.

\begin{table}[h]
\centering
\resizebox{1.00\linewidth}{!}{
\begin{tabular}{l|cc}
\toprule
\textbf{Method} & \textbf{w/o SoftMask} & \textbf{w/ SoftMask} \\
\midrule
LoRA $r=16$ & 72.18 & 72.12 \\
LoRA $r=32$ & 72.08 & 72.23 \\
LoRA $r=64$ & 71.83 & 72.40 \\
Full fine-tuning & 71.50 & 73.52 \\
\bottomrule
\end{tabular}}
\caption{Results on MTEB English with and without SoftMask.}
\label{table:train_gap}
\end{table}

\section{Conclusion}
In this paper, we propose Conan-embedding-v2, a new LLM trained from scratch and finetuned as a text embedder. We first address the data and training gaps between LLM and embedding models.
By leveraging pairs for LLM training, soft mask for embedding weakly-supervised training, cross-lingual retrieval dataset and dynamic hard negative mining for embedding supervised training, Conan-embedding-v2 achieves SOTA while maintaining a reasonable model size and inference speed.

Embedding models are crucial tools that empower fields like recommendation systems, text matching, and entity recognition.
We hope to inspire future research in embedding training methods and aim to explore more applications. 
In the future, we plan to continue updating our model to improve the performance and extend the capabilities in cross-modal retrieval.

\section*{Limitations}
\subsection*{A Cross lingual Retrieval Data Analysis}
To better understand the effectiveness and limitations of the cross-lingual retrieval dataset construction method proposed in Section~\ref{sec:cross_lingual}, we analyze the potential impact of language distribution within the dataset.

\paragraph{A.1 Proportion of Different Language Pairs.}
For cross-lingual retrieval, we employ T2Retrieval for Chinese-to-English retrieval and MSMARCO for multilingual retrieval by translating queries into 26 languages. For our translation process, we referenced the language distribution from the MTEB benchmarks to allocate language pairs, resulting in approximately 1 million pairs as shown in Table~\ref{tab:lang_dist}.

\begin{table}[h]
\centering
\caption{Language distribution for translation pairs.}
\label{tab:lang_dist}
\resizebox{1.00\linewidth}{!}{
\begin{tabular}{ll|ll}
\toprule
Language & Proportion & Language & Proportion \\
\midrule
English           & 25\% & Swedish      & 2\% \\
Chinese           & 12\% & Thai         & 2\% \\
Spanish           & 8\%  & Malay        & 2\% \\
French            & 6\%  & Turkish      & 2\% \\
Japanese          & 6\%  & Vietnamese   & 2\% \\
German            & 5\%  & Dutch        & 2\% \\
Russian           & 5\%  & Polish       & 2\% \\
Italian           & 4\%  & Hindi        & 2\% \\
Portuguese        & 4\%  & Khmer        & 1\% \\
Arabic            & 3\%  & Finnish      & 1\% \\
Korean            & 3\%  & Hebrew       & 1\% \\
Bengali           & 2\%  & Hungarian    & 1\% \\
Danish            & 2\%  & Norwegian    & 1\% \\

\bottomrule
\end{tabular}}
\end{table}

\paragraph{A.2 Performance on Specific Languages.}
Performance varies significantly across languages depending on the available resources. Table~\ref{tab:resource_perf} illustrates the performance metrics for languages with different resource levels evaluated on the MKQA dataset. Mid-resource languages, including Spanish, French, Japanese, German, Russian, Italian, and Portuguese, demonstrate better performance compared to low-resource languages. This performance gap likely stems from the disparity in available training data proportions.

Despite being a high-resource language, Chinese shows lower performance. This is likely due to the unique Chinese-English mapping relationship, which conflicts with MKQA's multilingual-to-English evaluation. In the future, we will focus on improving data processing for low-resource languageså and implementing balanced data sampling.

\begin{table}[h]
\centering
\caption{Performance by language resource level on MKQA.}
\label{tab:resource_perf}
\begin{tabular}{lcc}
\toprule
Resource & Proportion & Performance \\
\midrule
High-resource & 37\% & 70.6 \\
Mid-resource  & 45\% & 73.47 \\
Low-resource  & 18\% & 72.19 \\
\bottomrule
\end{tabular}
\end{table}

\paragraph{A.3 Potential Biases.}
The high proportion of English-Chinese data may lead to inflated performance metrics for cognate languages, potentially introducing biases across different language families and linguistic characteristics. We conducted an evaluation of performance across language families. Table~\ref{tab:family_perf} shows that Germanic, Slavic, and Romance languages (all Indo-European) exhibit strong performance. Notably, typologically distant languages like Arabic (65.2\%) and Korean (67.5\%) perform significantly lower, suggesting that linguistic similarity to English, rather than data volume, is the primary factor influencing model effectiveness. This highlights the challenge of achieving consistent performance across diverse language families.

\begin{table}[h]
\centering
\caption{Average performance and data share by language family.}
\label{tab:family_perf}
\begin{tabular}{lcc}
\toprule
Language Family & Avg. Score & Total Share \\
\midrule
Chinese   & 70.4 & 37\% \\
Germanic  & 74.6 & 36\% \\
Romance   & 73.9 & 22\% \\
Slavic    & 75.5 & 7\% \\
Arabic    & 65.2 & 3\% \\
Korean    & 67.5 & 3\% \\
Others    & 67.7 & 11\% \\
\bottomrule
\end{tabular}
\end{table}

\subsection*{B. Error Analysis}
Embedding models often struggle with numerical inconsistencies in semantically similar content. For example, when searching for "3 fairy tales", the model might give low similarity scores to content containing "5 fairy tales", even though the core content is relevant. This happens because embedding models treat numbers as regular tokens without understanding their quantitative relationship. Future improvements could include incorporating retrieval augmented generation to provide external numerical knowledge, and enriching training data with more numerical variations to enhance the model's understanding of quantitative relationships.
\bibliography{custom}

\appendix

\section{Implementation details}\label{sec:implementation}
The model is trained with a maximum input length of 32768 tokens. To enhance efficiency, mixed precision training and DeepSpeed ZERO-stage 1~\cite{rajbhandari2020zero} are utilized. 
For the LLM pre-training stage, we use AdamW~\cite{loshchilov2017decoupled} optimizer and learning rate of 1e-4, with 0.05 warmup ratio and 0.001 weight decay. The batch size is set to 256. The entire pre-training process employs 64 Ascend 910B GPUs and 219 hours.
For the LLM finetune stage, we use AdamW~\cite{loshchilov2017decoupled} optimizer and learning rate of 2e-5, with 0.02 warmup ratio and 0.001 weight decay. The batch size is set to 64. The entire pre-training process employs 16 Ascend 910B GPUs and 38 hours.
For the embedding weakly-supervised training stage, We used the same optimizer parameters and learning rate as in the pre-training phase. The batch size is set to 64. The entire pre-training process employs 16 Ascend 910B GPUs and 97 hours.
For the embedding supervised training stage, the MRL training representation dimensions are configured as 256, 512, 1024, 1536, 2048, 3072, 3584. The batch size is set to 4 for the retrieval task and 32 for the STS task. We sample 7 negatives for each query for retrieval task. We used the same optimizer parameters and learning rate as in the pre-training phase. The entire fine-tuning process employs 16 Ascend 910B GPUs and takes 13 hours.

\section{Data details}\label{sec:data_details}
In Section~\ref{sec:llm_train}, we have already introduced the datasets used during the LLM training phase. In Section~\ref{sec:train_data}, we have discussed the types of datasets used during the embedding weakly-supervised training and supervised training phases. 

\textbf{For Retrieval}: We utilized datasets such as TriviaQA~\cite{Joshi_Choi_Weld_Zettlemoyer_2017},
HotpotQA~\cite{Yang_Qi_Zhang_Bengio_Cohen_Salakhutdinov_Manning_2018},
NQ~\cite{Kwiatkowski_Palomaki_Redfield_Collins_Parikh_Alberti_Epstein_Polosukhin_Devlin_Lee_e2019},
MSMARCO~\cite{nguyen2016ms},
PubMedQA~\cite{Jin_Dhingra_Liu_Cohen_Lu_2019},
SQuAD~\cite{Rajpurkar_Zhang_Lopyrev_Liang_2016},
DuReader~\cite{He_Liu_Liu_Lyu_Zhao_Xiao_Liu_Wang_Wu_She_et_2018}, 
SimCSE~\cite{Gao_Yao_Chen_2021},
FEVER~\cite{Thorne_Vlachos_Christodoulopoulos_Mittal_2018}.

\textbf{For Reranking}: We used
StackOverFlow DupQuestions~\cite{liu2018linkso}
T2Ranking~\cite{Xie_Dong_Wang_Lv_Yao_Gan_Wu_Li_Li_Liu_et_2023},
CMedQAv2~\cite{zhang2018multi}.

\textbf{For Classification}: We used AmazonReviews~\cite{mcauley2013hidden}, AmazonCounterfactual~\cite{Rozenshtein_Kiryo_Kubota_Bollegala_2021}, Banking77~\cite{Casanueva_Temčinas_Gerz_Henderson_Vulić_2020}, Emotion~\cite{Saravia_Liu_Huang_Wu_Chen_2018}, TweetSentimentExtraction~\cite{tweet-sentiment-extraction}, MTOPIntent~\cite{li2020mtop}, IMDB~\cite{Maas_Daly_Pham_Huang_Ng_Potts_2011}, ToxicConversations~\cite{do2019jigsaw}, Tnews, Iflytek~\cite{Xu_Hu_Zhang_Li_2020}, Multilingualsentiments~\cite{McAuley_Leskovec_2013}.

\textbf{For Clustering}: We employed \{Arxiv/Biorxiv /Medrxiv/Reddit/StackExchange/Thunews/CSL\}-Clustering-{S2S/P2P}~\cite{muennighoff2022mteb,geigle2021tweac,Li_Sun_Zhang_2006,Li_Zhang_Zhao_Shen_Liu_Mao_Zhang}, TwentyNewsgroups~\cite{Lang_1995}.

\textbf{For STS}: We chose STS12~\cite{Agirre_Cer_Diab_Gonzalez2012}, STS22~\cite{chen2022semeval}, STS-Benchmark~\cite{Cer_Diab_Agirre_Lopez_2017}, AFQMC, QBQTC, Cmnli~\cite{Xu_Hu_Zhang_Li_2020} and Ocnli~\cite{Hu_Richardson_Xu_Li_Kübler_Moss_2020}.

For other languages, we leveraged the training data from Mr.Tydi~\cite{Zhang_Ma_Shi_Lin_2021} and MIRACL~\cite{zhang2023miracl}.

Table\ref{tab:weakly_data} and Table\ref{tab:finetune_data} shows that approximately 1.766 billion pairs were used during the weakly-supervised phase, and approximately 10.6 million pairs were used during the fine-tuning phase. The weakly-supervised training phase leverages a diverse collection of data sources, including News, Knowledge Base, Social Media, Web Page, Academic Paper, Community QA, and Instruction Datasets, as detailed in Table~\ref{tab:weakly_data}.
The supervised training phase, as shown in Table~\ref{tab:finetune_data}, focuses on specialized tasks such as Semantic Textual Similarity (STS), Contrastive Learning of Representations (CLR), Retrieval, and Classification.

\begin{table}[h]
    \centering
    \caption{Overview of the data sources used for embedding weakly-supervised training.}
    \resizebox{0.99\linewidth}{!}{
    \begin{tabular}{l|ccc}
      \toprule
      Categories & Data Format  & Numbers \\
      \midrule
      News & (title, content) & 620M \\
      Knowledge Base & (question, answer) & 106M \\
      Social Media & (title, content)  & 690M \\
      Web Page & (input, output) & 70M \\
      Academic Paper & (title, content)  & 50M  \\
      Community QA & (question, answer)  & 30M \\
      Instruction datasets & (prompt, response) & 200M \\
      \bottomrule
    \end{tabular}}
    \label{tab:weakly_data}
\end{table}

\begin{table}[h]
    \centering
    \caption{Overview of the data used for embedding supervised training.}
    \resizebox{\linewidth}{!}{
    \begin{tabular}{llcccc}
      \toprule
      Tasks & Data Format & Numbers \\
      \midrule
      STS & (sentence, sentence pairs) & 1.8M \\
      CLR & (text, pos text, neg text)  & 3.0M  \\
      Retrieval & (text, pos text, neg text)  & 3.0M  \\
      classification & (text, pos label, neg label)  & 2.8M  \\
      \bottomrule
    \end{tabular}}
    \label{tab:finetune_data}
\end{table}
\section{Soft Mask Function}\label{appendix:softmaskfunc}
For the $\alpha(t)$ mentioned in Sec~\ref{sec:soft_mask}, this section discusses three specific implementation functions: linear decay, quadratic decay (accelerating), and quadratic decay (decelerating). The specific formulas are as follows:

\begin{itemize}

\item Linear function: $\alpha(t) = \frac{t}{\tau}$
\item
    quadratic (accelerating): 
    $\alpha(t) = \left(\frac{t}{\tau}\right)^2$
\item
    quadratic (decelerating): 
    $\alpha(t) = 1 - \left(1 - \frac{t}{\tau}\right)^2$

\end{itemize}
where $t$ represents the current time step, and $\tau$ represents the total number of time steps.
We conducted comparative experiments on three different functions, exclusively utilizing the soft mask method and not the other two methods. As shown in Table~\ref{table:soft_func}, the Linear method yielded the best results, while the Decelerating method showed a decline in performance.
\begin{table}
    \centering
    \begin{tabular}{ccccccc}
    \toprule
    \textbf{Function}  & \textbf{Multi} & \textbf{Eng} & \textbf{Zh} \\
    \midrule
    Linear  & 61.73 & 70.41 &  70.99 \\
    Accelerating & 61.50 & 70.51 &  70.81 \\
    Decelerating & 61.43 & 70.01 &  70.37 \\
    \bottomrule
    \end{tabular}
    \caption{Results of different soft mask functions.}
    \label{table:soft_func}
\end{table}

\begin{table*}[ht]
    \centering
    \resizebox{\textwidth}{!}{
    \begin{tabular}{lccccccccc}
    \toprule
    & \textbf{BM25} & \textbf{mDPR} & \textbf{mContriever} & \textbf{Multilingual-E5-large} & \textbf{e5-mistral-7b-instruct} & \textbf{text-embedding-v3} & \textbf{M3-embedding} & \textbf{Conan-embedding-v2} \\
    \midrule
    ar      & 13.4 & 33.8 & 43.8 & 59.7 & 47.6 & 55.1 & 63.0 & 65.2 \\
    da      & 36.2 & 55.7 & 63.3 & 71.7 & 72.3 & 67.6 & 72.0 & 73.1 \\
    de      & 23.3 & 53.2 & 60.2 & 71.2 & 70.8 & 67.6 & 70.4 & 72.8 \\
    es      & 29.8 & 55.4 & 62.3 & 70.8 & 71.6 & 68.0 & 70.7 & 73.2 \\
    fi      & 33.2 & 42.8 & 58.7 & 67.7 & 63.6 & 65.5 & 68.9 & 71.6 \\
    fr      & 30.3 & 56.5 & 62.6 & 69.5 & 72.7 & 68.2 & 70.8 & 73.5 \\
    he      & 16.1 & 34.0 & 50.5 & 61.4 & 32.4 & 46.3 & 64.6 & 66.7 \\
    hu      & 26.1 & 46.1 & 57.1 & 68.0 & 68.3 & 64.0 & 67.9 & 70.2 \\
    it      & 31.5 & 53.8 & 62.0 & 71.2 & 71.3 & 67.6 & 70.3 & 73.9 \\
    ja      & 14.5 & 46.3 & 50.7 & 63.1 & 57.6 & 64.2 & 67.9 & 71.8 \\
    km      & 20.7 & 20.6 & 18.7 & 18.3 & 23.3 & 25.7 & 59.5 & 62.4 \\
    ko      & 18.3 & 36.8 & 44.9 & 58.9 & 49.4 & 53.9 & 63.3 & 67.5 \\
    ms      & 42.3 & 53.8 & 63.7 & 70.2 & 71.1 & 66.1 & 72.3 & 78.4 \\
    nl      & 42.5 & 56.9 & 63.9 & 73.0 & 74.5 & 68.8 & 72.3 & 75.6 \\
    no      & 38.5 & 55.2 & 63.0 & 71.1 & 70.8 & 67.0 & 71.6 & 76.9 \\
    pl      & 28.7 & 50.4 & 60.9 & 70.5 & 71.5 & 66.1 & 70.4 & 76.7 \\
    pt      & 31.8 & 52.5 & 61.0 & 66.8 & 71.6 & 67.7 & 70.6 & 74.8 \\
    ru      & 21.8 & 49.8 & 57.9 & 70.6 & 68.7 & 65.1 & 70.0 & 74.3 \\
    sv      & 41.1 & 54.9 & 62.7 & 72.0 & 73.3 & 67.8 & 71.5 & 74.8 \\
    th      & 28.4 & 40.9 & 54.4 & 69.7 & 57.1 & 55.2 & 70.8 & 75.9 \\
    tr      & 33.5 & 45.5 & 59.9 & 67.3 & 65.5 & 64.9 & 69.6 & 75.8 \\
    vi      & 33.6 & 51.3 & 59.9 & 68.7 & 62.3 & 63.5 & 70.9 & 73.0 \\
    zh\_cn  & 19.4 & 50.1 & 55.9 & 44.3 & 61.2 & 62.7 & 67.3 & 70.4 \\
    zh\_hk  & 23.9 & 50.2 & 55.5 & 46.4 & 55.9 & 61.4 & 66.7 & 71.8 \\
    zh\_tw  & 22.5 & 50.6 & 55.2 & 45.9 & 56.5 & 61.6 & 65.6 & 69.7 \\
    \midrule
    Avg    & 28.1 & 47.9 & 56.3 & 63.5 & 62.4 & 62.1 & 68.8 & 72.4 \\
    \bottomrule
    \end{tabular}}
    \caption{Recall@20 on MKQA dataset for cross-lingual retrieval in all 25 languages.}
    \label{tab:mkqa_recall@20_results}
\end{table*}
\section{More Results}
\subsection{MKQA Results}\label{appendix:mkqa}
In this section, we present the results for all languages on the MKQA benchmark. As shown in Table~\ref{tab:mkqa_recall@20_results}, Conan-embedding-v2 outperforms all baselines on average.

\subsection{MTEB Results}\label{appendix:mteb}
In this section, we present additional evaluation results on the MTEB English benchmark and MTEB Chinese benchmarks. As shown in Table~\ref{table:mteb_en} and Table~\ref{tab:mteb_zh}, Conan-embedding-v2 outperforms all baselines on average.

\begin{table*}[ht]
\centering
\resizebox{\textwidth}{!}{%
\begin{tabular}{lcccccccccccc}
\toprule
\textbf{} & \textbf{Bge-multilingual-gemma2} & \textbf{Gte-Qwen2-7B-instruct} & \textbf{SFR-Embedding-2R} & \textbf{Stella-en1.5B-v5} & \textbf{bge-en-icl} & \textbf{Conan-embedding-v2}\\
\midrule
ArguAna & 77.37 & 64.27 & 62.34 & 65.27 & 82.76 & 88.18 \\
ClimateFEVER & 39.47 & 45.88 & 34.43 & 46.11 & 45.35 & 44.45  \\
CQADupStack & 47.94 & 46.43 & 46.11 & 47.75 & 47.23 & 52.11  \\
DBPedia & 51.37 & 52.42 & 51.21 & 52.28 & 50.42 & 56.33  \\
FEVER & 90.38 & 95.11 & 92.16 & 94.83 & 91.96 & 92.52  \\
FiQA2018 & 60.04 & 62.03 & 61.17 & 60.48 & 58.77 & 62.16  \\
HotpotQA & 83.26 & 73.08 & 81.36 & 76.67 & 84.98 & 83.36  \\
MSMARCO & 45.71 & 45.92 & 42.18 & 45.22 & 46.72 & 52.38  \\
NFCorpus & 38.11 & 40.6 & 41.34 & 42 & 40.69 & 42.09  \\
Natural Questions & 71.45 & 67.73 & 73.96 & 71.8 & 73.85 & 82.81  \\
QuoraRetrieval & 90.04 & 90.09 & 89.58 & 90.03 & 91.02 & 90.58 \\
SCIDOCS & 26.93 & 28.91 & 24.87 & 26.64 & 25.25 & 30.21  \\
SciFact & 72.05 & 79.06 & 85.91 & 80.99 & 78.33 & 87.60  \\
Touche2020 & 30.26 & 30.57 & 28.18 & 29.94 & 29.67 & 31.09  \\
TREC-COVID & 64.27 & 82.26 & 87.28 & 85.98 & 78.11 & 93.87  \\

BIOSSES & 85.74 & 81.37 & 87.6 & 83.11 & 86.35 & 84.78  \\
SICK-R & 82.66 & 79.28 & 77.01 & 82.99 & 83.7 & 81.91  \\
STS12 & 77.71 & 79.55 & 75.67 & 80.09 & 77.73 & 84.07  \\
STS13 & 87.45 & 88.83 & 82.94 & 86.09 & 85.98 & 86.7 \\
STS14 & 83.48 & 85.73 & 78.43 & 87.32 & 82.94 & 83.18  \\
STS15 & 87.63 & 88.54 & 85.82 & 89.13 & 86.54 & 86.54 \\
STS16 & 86.49 & 85.84 & 87.15 & 86.54 & 87.24 & 87.52 \\
STS17 & 91.18 & 88.93 & 88.9 & 91.05 & 91.82 & 89.09  \\
STS22 & 69.02 & 66.88 & 67.1 & 68.01 & 68.08 & 69.3  \\
STSBenchmark & 87.25 & 83.63 & 88.23 & 88.92 & 86.14 & 87.01 \\

SummEval & 31.2 & 31.35 & 31.4 & 30.75 & 30.70 & 30.64 \\

SprintDuplicateQuestions & 79.32 & 97.62 & 97.61 & 97.05 & 95.04 & 94.99 \\
TwitterSemEval2015 & 79.64 & 77.88 & 80.58 & 78.54 & 78.73 & 80.34 \\
TwitterURLCorpus & 86.95 & 86.59 & 88.03 & 87.58 & 87.19 & 89.38  \\

AmazonCounterfactual & 98.49 & 98.87 & 97.88 & 97.89 & 95.12 & 97.12 \\
AmazonPolarity & 96.9 & 97.31 & 97.1 & 96.86 & 97.14 & 98.91 \\
AmazonReviews & 62.56 & 61.04 & 59.36 & 61.28 & 61.47 & 66.01 \\
Banking77 & 92.53 & 90.2 & 90.41 & 90.41 & 90.34 & 91.05 \\
Emotion & 92.97 & 79.45 & 93.37 & 84.29 & 93.31 & 93.68 \\
Imdb & 96.66 & 96.8 & 96.7 & 96.8 & 96.7 & 96.9 \\
MassiveIntent & 82.05 & 85.7 & 85.85 & 85.83 & 82.26 & 88.71 \\
MassiveScenario & 84.4 & 89.97 & 90.61 & 90.21 & 83.92 & 90.1 \\
MTOPDomain & 98.61 & 98.04 & 98.1 & 98.2 & 96.51 & 95.76 \\
MTOPIntent & 95.51 & 91.88 & 91.3 & 92.78 & 93.56 & 96.97 \\
ToxicConversations & 85.12 & 91.14 & 88.75 & 93.16 & 92.77 & 93.08 \\
TweetSentimentExtraction & 78.58 & 79.7 & 74.84 & 78.3 & 80.6 & 85.03 \\

Arxiv-P2P & 54.91 & 54.46 & 54.02 & 55.44 & 54.42 & 56.31 \\
Arxiv-S2S & 50.28 & 51.74 & 48.82 & 51.44 & 49.59 & 57.03 \\
Biorxiv-P2P & 52.64 & 50.09 & 50.76 & 50.68 & 52.32 & 52.32 \\
Biorxiv-S2S & 49.2 & 46.56 & 47.67 & 48.67 & 44.36 & 48.39 \\
Medrxiv-P2P & 45.81 & 46.23 & 46.66 & 46.8 & 46.13 & 46.19  \\
Medrxiv-S2S & 44.11 & 44.18 & 44.65 & 44.65 & 41.36 & 46.58  \\
Reddit & 56.03 & 73.55 & 62.92 & 72.86 & 71.2 & 72.32 \\
Reddit-P2P & 65.83 & 74.13 & 72.74 & 75.27 & 72.17 & 76.15  \\
StackExchange & 66.21 & 79.86 & 76.48 & 80.29 & 81.29 & 82.13 \\
StackExchange-P2P & 45.74 & 49.4 & 48.29 & 49.57 & 45.53 & 53.64 \\
TwentyNewsgroups & 70.44 & 53.91 & 66.42 & 61.43 & 68.51 & 64.17 \\
AskUbuntuDupQuestions & 64.59 & 67.58 & 66.71 & 67.33 & 64.8 & 67.46 \\
MindSmallRank & 31.79 & 33.36 & 31.26 & 33.05 & 30.6 & 33.28  \\
SciDocsRR & 87.6 & 89.09 & 87.29 & 89.2 & 86.9 & 88.94  \\
StackOverflowDupQuestions & 54.9 & 55.06 & 55.32 & 55.25 & 56.32 & 56.28 \\
\midrule
MTEB Average (56) & 69.88 & 70.24 & 70.31 & 71.19 & 71.24 & 73.09 \\
\bottomrule
\end{tabular}}
\caption{MTEB English benchmark.}
\label{table:mteb_en}
\end{table*}

\begin{table*}
    \centering
    \resizebox{\textwidth}{!}{%
    \begin{tabular}{lcccccccccccc}
    \toprule
    \multirow{2}{*}{\textbf{}}
    & \textbf{e5-mistral} & \textbf{gte-Qwen2} & \textbf{xiaobu-} & \textbf{Conan-} & \textbf{bge-multilingual-} & \textbf{gte-Qwen2} & \textbf{Conan-} \\
    & \textbf{-7b-instruct} & \textbf{-7B-instruct} & \textbf{embedding-v2} & \textbf{embedding-v1} & \textbf{gemma2} & \textbf{-1.5B-instruct}& \textbf{embedding-v2} & \\
    \midrule
    
    CmedqaRetrieval & 34.23 & 48.69 & 47.14 & 47.61 & 42.21 & 46.97 & 45.32 \\
    CovidRetrieval & 73.11 & 81.04 & 89.40 & 92.35 & 77.46 & 80.79 & 79.88 \\
    DuRetrieval & 87.04 & 87.44 & 89.44 & 88.53 & 90.46 & 89.40 & 88.72 \\
    EcomRetrieval & 45.95 & 71.15 & 70.50 & 70.99 & 69.30 & 62.51 & 68.12 \\
    MMarcoRetrieval & 74.84 & 85.16 & 82.19 & 82.25 & 84.70 & 83.01 & 83.45 \\
    MedicalRetrieval & 52.83 & 65.59 & 68.19 & 67.94 & 62.02 & 58.65 & 62.56 \\
    T2Retrieval & 80.68 & 87.73 & 85.01 & 83.31 & 86.26 & 85.47 & 84.92 \\
    VideoRetrieval & 45.34 & 78.84 & 80.09 & 80.40 & 77.40 & 68.11 & 76.55 \\
	
    Ocnli & 80.21 & 90.18 & 92.84 & 92.54 & 86.22 & 90.13 & 92.74 \\
    Cmnli & 72.19 & 87.48 & 91.87 & 91.66 & 86.91 & 86.67 & 89.90 \\
    
    AmazonReviews & 47.6 & 53.55 & 50.07 & 50.31 & 54.34 & 52.95 & 53.81 \\
    MassiveIntent & 72.46 & 81.09 & 77.45 & 78.14 & 78.19 & 76.25 & 80.51 \\
    MassiveScenario & 76.4 & 85.74 & 85.3 & 86.2 & 82.58 & 77.26 & 86.45 \\
    IFlyTek & 48.65 & 54.52 & 51.76 & 51.94 & 49.94 & 44.85 & 50.32 \\
    JDReview & 84.69 & 86.51 & 89.08 & 90.32 & 88.91 & 85.82 & 90.09 \\
    MultilingualSentiment & 74.64 & 76.88 & 79.45 & 78.58 & 78.91 & 77.42 & 80.17 \\
    OnlineShopping & 92.56 & 94.30 & 94.90 & 95.07 & 94.59 & 93.50 & 94.19 \\
    TNews & 50.58 & 52.97 & 54.64 & 55.03 & 50.26 & 49.95 & 58.21 \\
    Waimai & 87.79 & 89.47 & 89.34 & 89.70 & 89.26 & 86.63 & 88.45 \\
    
    CMedQAv1-reranking & 76.82 & 88.20 & 90.96 & 91.39 & 84.62 & 88.16 & 91.81 \\
    CMedQAv2-reranking & 77.59 & 89.31 & 90.41 & 89.72 & 85.60 & 88.12 & 89.45 \\
    MMarcoReranking & 24.21 & 31.65 & 39.91 & 41.58 & 35.43 & 29.14 & 41.59 \\
    T2Reranking & 66.90 & 67.80 & 69.03 & 68.36 & 67.48 & 67.43 & 71.91 \\
    
    AFQMC & 38.99 & 72.25 & 60.96 & 60.66 & 47.17 & 58.42 & 60.32 \\
    ATEC & 43.58 & 62.62 & 58.81 & 58.64 & 50.75 & 55.65 & 59.23 \\
    BQ & 54.67 & 81.25 & 75.08 & 74.51 & 62.02 & 73.85 & 74.63 \\
    LCQMC & 75.48 & 73.81 & 79.82 & 79.45 & 75.95 & 75.39 & 80.66 \\
    PAWSX & 16.81 & 54.06 & 47.42 & 46.60 & 30.57 & 42.46 & 45.17 \\
    QBQTC & 31.80 & 31.37 & 45.14 & 44.58 & 38.98 & 35.15 & 43.98 \\
    STSB & 84.77 & 83.88 & 82.05 & 81.24 & 80.87 & 79.4 & 81.15 \\
    STS22 & 63.4 & 65.77 & 66.96 & 67.73 & 68.68 & 67.4 & 68.78 \\
    
    CLSClusteringP2P & 44.42 & 47.07 & 60.42 & 60.64 & 54.65 & 45.21 & 64.48 \\
    CLSClusteringS2S & 42.58 & 45.99 & 49.54 & 52.65 & 63.68 & 42.50 & 62.83 \\
    ThuNewsClusteringP2P & 64.68 & 86.08 & 78.76 & 77.84 & 64.32 & 68.24 & 76.11 \\
    ThuNewsClusteringS2S & 57.53 & 85.11 & 71.96 & 74.20 & 54.57 & 62.50 & 73.59 \\
    \midrule
    MTEB Average (35) & 60.89 & 71.94 & 72.43 & 72.62 & 68.44 & 67.75 & 72.83 \\
    \bottomrule
    \end{tabular}}
    \caption{MTEB Chinese benchmark.}
    \label{tab:mteb_zh}
    \end{table*}

\end{document}